\documentclass[runningheads]{llncs}
\usepackage[T1]{fontenc}
\usepackage{graphicx}
\usepackage{hyperref}
\usepackage{url}
\usepackage{amsfonts}
\usepackage{caption}
\usepackage{subcaption}
\usepackage{amsmath}
\usepackage{physics}
\usepackage{tikz}
\usetikzlibrary{quantikz}
\newcommand{\mty}{\underline{\hspace{0.25cm}}}
\usepackage[symbol]{footmisc}

\begin{document}
\title{Reinforcement learning for Quantum Tiq-Taq-Toe}
%
%
\author{Catalin Dinu \and
Thomas Moerland
}
\authorrunning{Dinu et al.}
%
\institute{Leiden Institute of Advanced Computer Science, Leiden University, The Netherlands }
\maketitle              
\section{Introduction}
Different quantum adaptations of Tic-Tac-Toe exist, such as Quantum Tic-Tac-Toe by Goff \cite{goff2006quantum}. In this work, we use the rules and implementations provided by TiqTaqToe.com \cite{tiqtaqtoe} and Quantumlib \cite{git}, which represent the game directly using quantum states. Unlike Goff's quantum-inspired formulation, the version considered here uses qutrit-based quantum states and is compatible with quantum-hardware implementations. A detailed explanation of the rules is provided in Appendix A. Despite its popularity, no reinforcement learning (RL) methods have been applied to Quantum Tiq-Taq-Toe. Although there has been some research on Quantum Chess \cite{youvan2024sequential,cantwell2019quantum}, this game is significantly more complex in terms of computation and analysis. Therefore, we study the combination of quantum computing and reinforcement learning in Quantum Tiq-Taq-Toe (code for our work can be found \cite{work}), which may serve as an accessible testbed for the integration of both fields.

\section{Methodology}
Quantum games are challenging to represent classically due to their inherent partial observability and the potential for exponential state complexity. In Quantum Tiq-Taq-Toe, states are observed through Measurement (a 3x3 matrix of state probabilities) and Move History (a 9x9 matrix of entanglement relations), making strategy complex as each move can collapse the quantum state.

Our study examines two versions of Quantum Tiq-Taq-Toe from the {\it quantumlib} repository \cite{git}. The first version restricts entanglement moves to include at least one empty cell, blending traditional rules with quantum mechanics. The second version lifts these restrictions, allowing more diverse quantum states and interactions, thereby increasing strategic depth.
\section{Results}
We conducted a comparative analysis of self-play PPO \cite{schulman2017proximal,tang2020implementing,liu2021self} agents in Quantum Tiq-Taq-Toe, exploring their performance with access to both measurement matrices and historical entanglement records (M\&H agent), as well as with access to only the measurement matrix (M) or historical entanglement record (H).

\begin{figure}[ht]
\caption{RL results on Quantum Tiq-Tac-Toe Version 1}
\centering
\begin{subfigure}{.48\textwidth}
  \centering
  \includegraphics[width=0.7\linewidth]{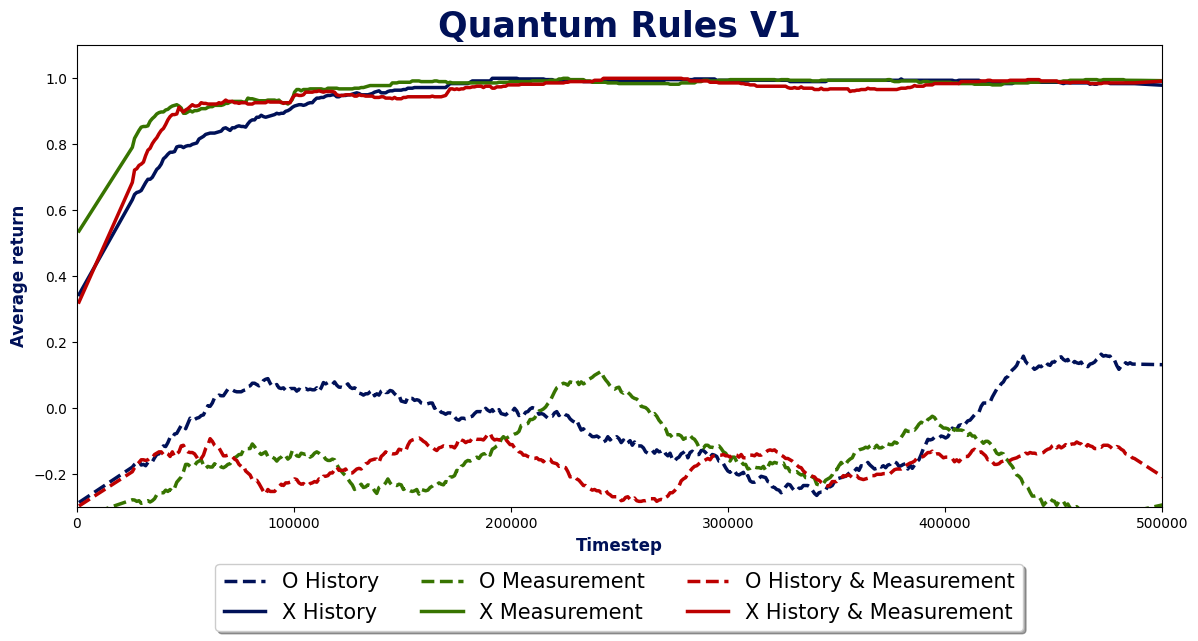}
  \caption{Average reward on 100 games during the training of different agents }
  \label{fig:reward_v1}
\end{subfigure}%
\hfill
\begin{subfigure}{.48\textwidth}
  \centering
  \includegraphics[width=0.7\linewidth]{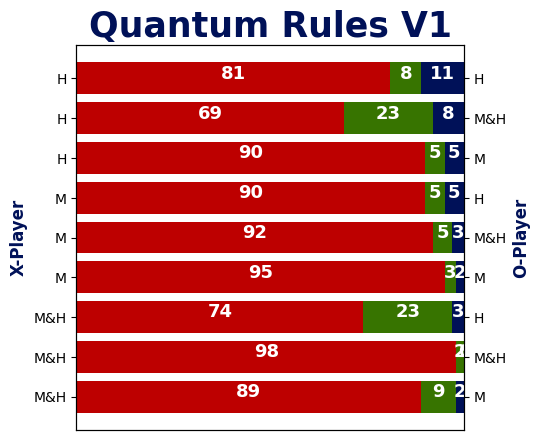}
  \caption{Pitting best agents: Red $\xrightarrow[]{}$ X-Wins, Blue $\xrightarrow[]{}$ O-Wins, Green $\xrightarrow[]{}$ Draws  }
  \label{fig:results_v1}
\end{subfigure}

\label{fig:qrv1}
\end{figure}

For the first set of rules, which imposes constraints on entanglement moves, we observe a tendency for the first player to gain an advantage (Fig. \ref{fig:qrv1}). This advantage is noticeable despite inherent randomness in the game, which prevents guaranteed wins, and therefore suggests the presence of discernible winning strategies.

\begin{figure}[t]
\centering
\caption{RL results on Quantum Tiq-Tac-Toe Version 3}
\begin{subfigure}{.48\textwidth}
  \centering
  \includegraphics[width=0.7\linewidth]{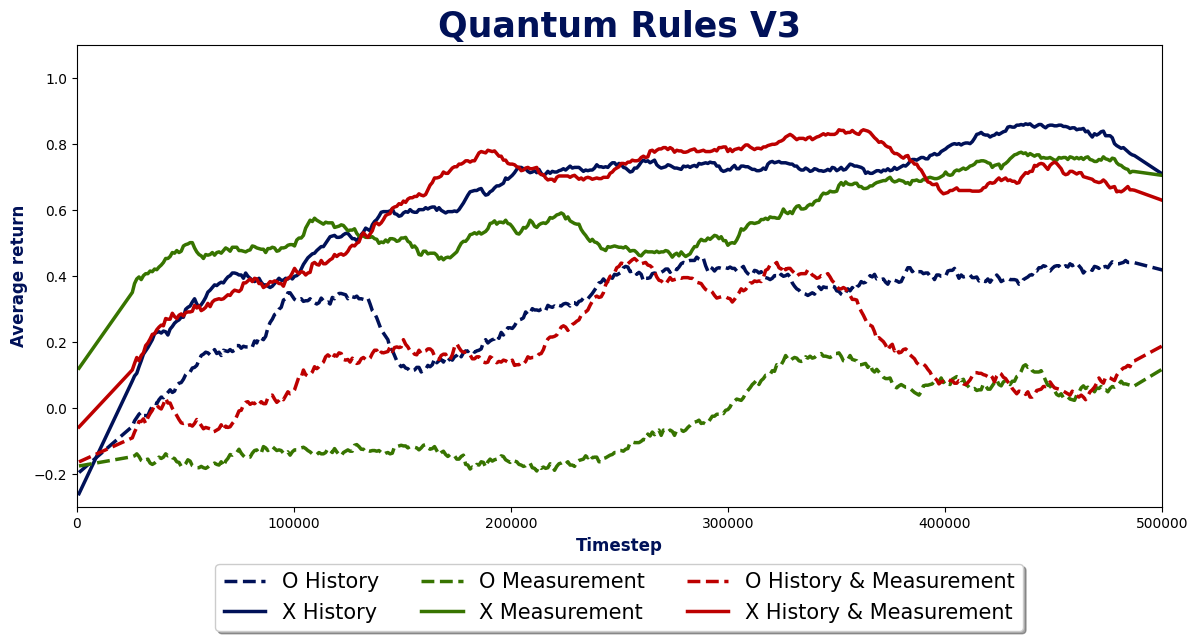}
  \caption{Average reward on 100 games during the training of different agents}
  \label{fig:reward_v3}
\end{subfigure}%
\hfill
\begin{subfigure}{.48\textwidth}
  \centering
  \includegraphics[width=0.7\linewidth]{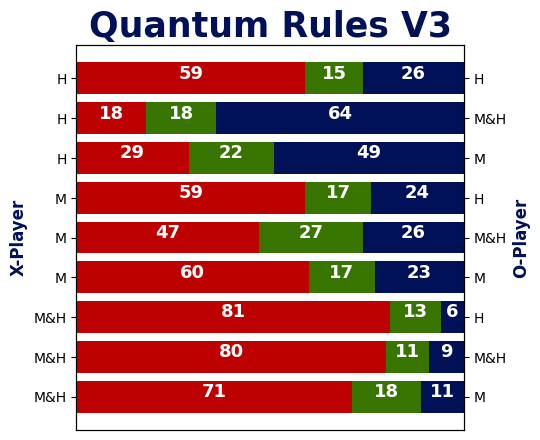}
  \caption{Pitting best agents: Red $\xrightarrow[]{}$ X-Wins, Blue $\xrightarrow[]{}$ O-Wins, Green $\xrightarrow[]{}$ Draws }
  \label{fig:results_V3}
\end{subfigure}
\label{fig:qrv3}
\end{figure}

For the third set of rules, which allows for triple entanglement, the combined state of measurement matrix and historical entanglement yields optimal performance based on the pitting results (Fig. \ref{fig:qrv3}). This integrated approach enables agents to utilize real-time state probabilities and insights from past game interactions, leading to more equitable outcomes between players. It also underscores the importance of comprehensive information in strongly partially observable quantum environments.
\section{Discussion}
Most quantum problems require both precise control and mitigation of partial observability, making machine learning particularly suitable. Indeed, RL has already shown promise in fields like quantum error correction \cite{andreasson2019quantum,nautrup2019optimizing}. We identify Quantum Tiq-Taq-Toe, with its various subtypes, as an accessible testbed for the development of RL methods in the quantum setting. Future work could investigate other methods to mitigate partial observability, such as the use of state windowing \cite{lin1993reinforcement}, recurrent neural networks \cite{hochreiter1997long}, recurrent state space models \cite{gu2023mamba}, or transformers \cite{vaswani2017attention}.

%
%
\bibliographystyle{splncs04}
\bibliography{ref}

\pagebreak
\section*{Appendix}
\subsection*{A. Environment} \label{app:env}

Quantum Tiq-Taq-Toe \cite{chiofalo2022games} is an altered variant of the classic Tic-Tac-Toe game. In the traditional game, each cell on a 3x3 board can be empty, X, or O. In the quantum version, each cell is a qutrit, existing in a superposition of three quantum states: empty ($\ket{\mty}$), X ($\ket{X}$), or O ($\ket{O}$). Given the full board state ($\ket{\eta}$), the probability of collapsing to a specific state $\ket{c_1c_2...c_8c_9}$ ($c_i \in \{\mty,X,O\}$) is:
\begin{align*}
    \ket{\eta}  &= \sum_{\phi \in \{\mty,X,O\}^9} \alpha_{\phi} \ket{\phi},\,
    \sum_{\phi \in \{\mty,X,O\}^9} \alpha_{\phi} ^2 =1 \\ 
    \mathbb{P}(\eta = c_1c_2...c_8c_9) &= ||\bra{c_1c_2...c_8c_9}\ket{\eta}||^2 = \alpha_{c_1c_2...c_8c_9} ^2
\end{align*}
where $\ket{\eta}$ is the quantum state of the system, that can be express as a linear combination of all possible classical states $(\ket{\phi})$ with $\alpha_{\phi}^2$ being the probability to observe the state $\ket{\phi}$.
\subsubsection{Action Space} 
\begin{figure}[h]
    \centering
    \includegraphics[width=0.5\linewidth]{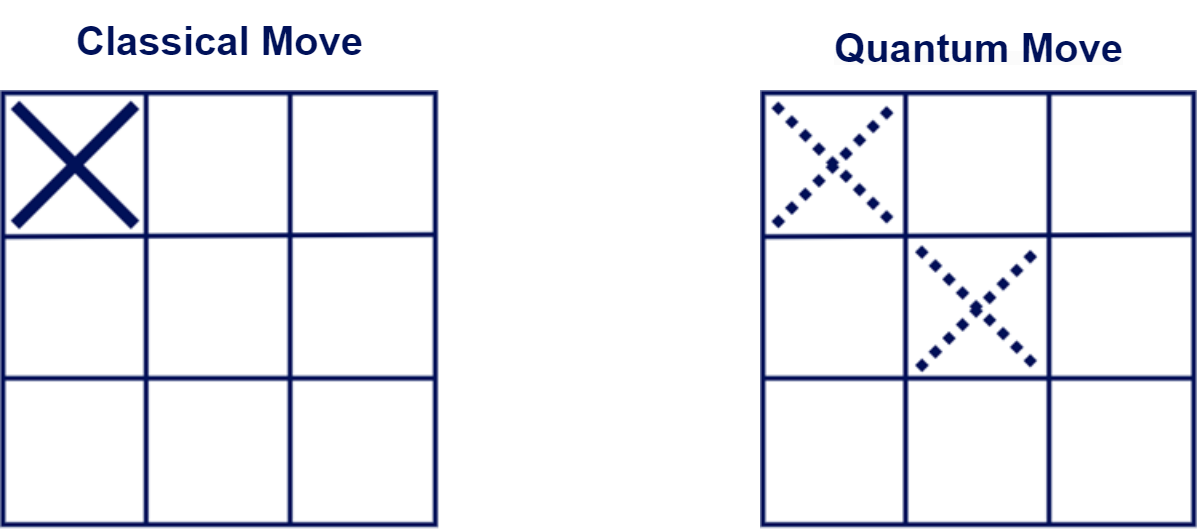}
    \caption{Most simple classical/quantum moves allowed during the game}
    \label{fig:moves_example}
\end{figure}

In classical Tic-Tac-Toe, a move changes an empty cell to X or O (left of Figure \ref{fig:moves_example}). In the quantum version, these moves are $X_{NOT}\ket{\mty} \xrightarrow[]{} \ket{X}$ and $O_{NOT}\ket{\mty} \xrightarrow[]{} \ket{O}$.

Additionally, quantum moves involve entangled pairs of cells. These moves create two possible states: one with X/O in the first cell and another with X/O in the second cell, leading to complex quantum states. The simplest case entangles two empty cells with X/O, resulting in a 50\% probability of X/O appearing in either cell (right of Fig. \ref{fig:moves_example}).
\begin{figure}[h]
    \centering
    \includegraphics[width=0.5\linewidth]{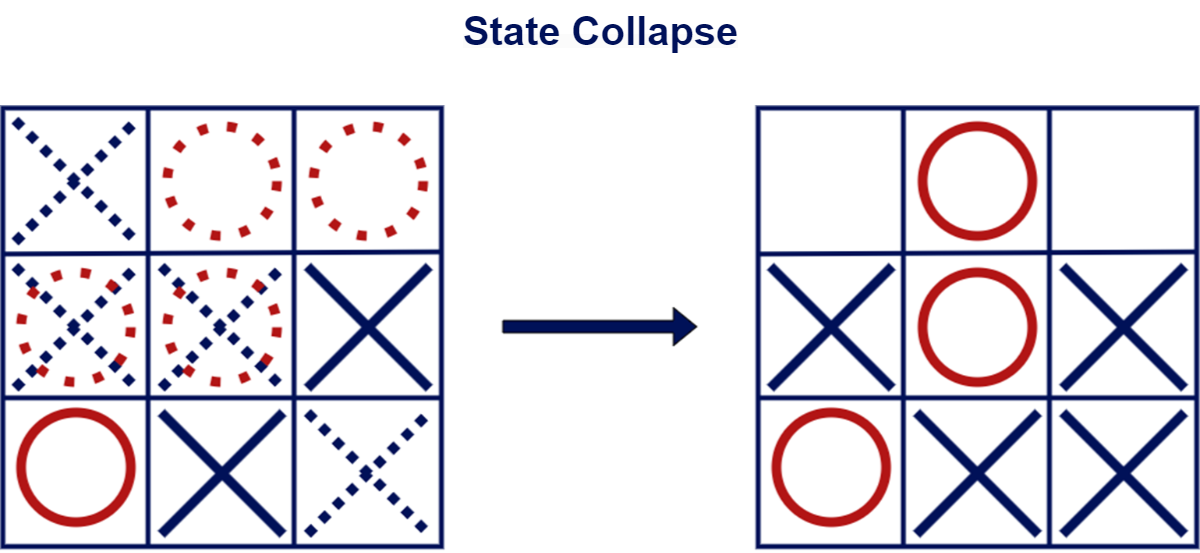}
    \caption{State collapsing after filling all the cells with anything (quantum/classical moves)}
    \label{fig:collapse}
\end{figure}
\subsubsection{State Collapsing} A pivotal phenomenon in the quantum game is State Collapsing, as illustrated in Fig.\ref{fig:collapse}. This occurrence occurs when the game board becomes saturated with moves, utilizing both quantum and classical moves that impact all cells on the board. Upon the state collapsing, a specific state is selected from the multitude of possible states. This selection is determined by the probability distribution outlined by the existing quantum state ($\ket{\eta} = \sum_{\phi \in \{\mty,X,O\}^9} \alpha_{\phi} \ket{\phi}$).

\begin{figure}[h]
    \centering
    \includegraphics[width=0.5\linewidth]{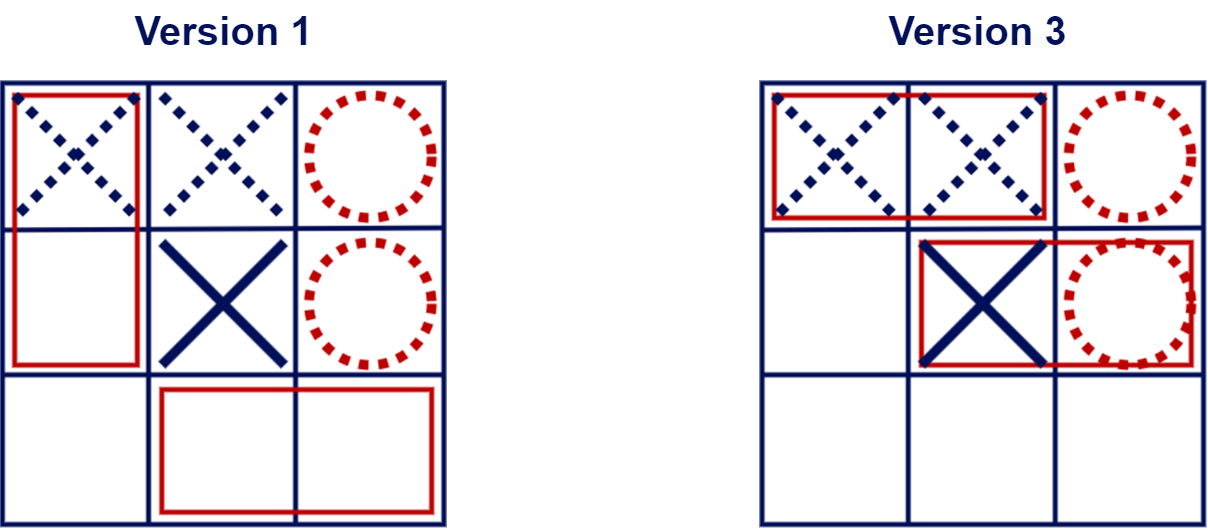}
    \caption{Available moves considering the two sets of game rules}
    \label{fig:gamerules}
\end{figure}

In addition, we distinguish two sets of game rules (Fig.\ref{fig:gamerules}) that affect the game play:
\begin{itemize}
    \item Version 1 (V1) - reduce the list of available entanglements moves only to pairs of cells that contain at least one free cell (we consider a free cell as a cell that was not used for any previous moves).
    \item Version 3 (V3) - any combination of two cells is a valid entanglement move.
\end{itemize}

\subsection*{B. Observation Space}

A paramount challenge in this game revolves around effectively representing quantum information in a classical format to facilitate an Agent's learning process. The most straightforward method involves classically storing the quantum state and simulating the game. However, this approach is deemed unreliable due to the impracticality of saving a 9-qutrit quantum state, which requires complex numbers. This not only contradicts the essence of a quantum game but also poses a significant computational burden.\\
To address this challenge, two classical pieces of information are explored in this report: Measurements and Moves History. These representations offer a more manageable way to capture and convey quantum aspects within the framework of the game, enabling effective learning for an Agent.
\subsubsection{Measurements}
A piece of essential information that can help an Agent learn to play the game would be the probability of each cell being in either \mty/X/O state. However, to compute the real values of those would imply the access to the quantum state. To overcome this, we can estimate those probabilities. Given the quantum state of the game board $\ket{\eta}$, we can simulate the state collapsing a number of times (N) and estimate the probabilities $\widehat{\mathbb{P}(c_i = \mty)}$, $\widehat{\mathbb{P}(c_i = X)}$and $\widehat{\mathbb{P}(c_i = O)}$ as the number of appearances of \mty/X/O on each cell divided by N. The relation between the estimates and real values is: $\widehat{\mathbb{P}(c_i = \mty)} = \mathbb{P}(c_i = \mty) + \mathcal{N}(0,\frac{1}{N})$, $\widehat{\mathbb{P}(c_i = X)} = \mathbb{P}(c_i = X) + \mathcal{N}(0,\frac{1}{N})$ and $\widehat{\mathbb{P}(c_i = O)} = \mathbb{P}(c_i = O) + \mathcal{N}(0,\frac{1}{N})$

So, these estimations provide a practical means for an Agent to learn and make decisions based on approximated probabilities, offering a computationally feasible approach in the absence of direct access to the precise quantum state.

\subsubsection{Moves History}
An additional informative resource for an agent's learning process involves maintaining a history of past moves. This data is structured using two matrices, one for X and one for O, each with dimensions of $9\times9$ ($MH^{X}$ and $MH^{O}$). The matrices are defined as follows:
\begin{itemize}
    \item $MH^{X/O}_{i,j}$: Represents the number of moves entangling $c_i$ and $c_j$ using X/O, where $i, j \in \{1, ..., 9\}$ and $i \neq j$.
    \item $MH^{X/O}_{i,i}$: Indicates the number of classical moves using X/O on $c_i$, where $i \in \{1, ..., 9\}$.
\end{itemize}
This fixed-dimension representation efficiently captures the historical moves in a structured manner, providing valuable information for the agent's learning process.

\end{document}